\documentclass{article}

\usepackage[numbers]{natbib}
\usepackage[preprint,nonatbib]{neurips_2024}

\usepackage[utf8]{inputenc} % allow utf-8 input
\usepackage[T1]{fontenc}    % use 8-bit T1 fonts
\usepackage{hyperref}       % hyperlinks
\usepackage{url}            % simple URL typesetting
\usepackage{booktabs}       % professional-quality tables
\usepackage{amsfonts}       % blackboard math symbols
\usepackage{nicefrac}       % compact symbols for 1/2, etc.
\usepackage{microtype}      % microtypography
\usepackage{xcolor}         % colors

\usepackage{microtype}
\usepackage{graphicx}
\usepackage{subcaption}
\usepackage{booktabs} % for professional tables
\usepackage{algorithm}
\usepackage{algorithmic}
\usepackage{hyperref}

\usepackage{amsmath}
\usepackage{amssymb}
\usepackage{mathtools}
\usepackage{amsthm}

\usepackage{enumitem}

\usepackage[capitalize,noabbrev]{cleveref}

\theoremstyle{plain}

\theoremstyle{definition}

\theoremstyle{remark}

\newcommand{\parag}[1]{\textbf{#1}}

\newcounter{displaybox}
\usepackage{tcolorbox}
\newenvironment{displaybox}[2][]{%
    \refstepcounter{displaybox} % Increment the box counter
    \begin{tcolorbox}[colback=white!95!gray,colframe=gray!25!black,arc=0mm,title=Box \thedisplaybox: #2,#1]
}{%
    \end{tcolorbox}
}

\usepackage{xspace}
\newcommand{\emphcol}[1]{\emph{\textcolor{red}{#1}}}
\newcommand{\name}{LaTable\xspace}
\newcommand{\Enc}{f_{\mathrm{LM}}}
\newcommand{\sk}{^{(k)}}
\newcommand{\catindices}{\mathcal{I}_{cat}\sk}
\newcommand{\bbR}{\mathbb{R}}

\title{LaTable: Towards Large Tabular Models}

\author{%
  Boris van Breugel \quad Jonathan Crabbé \quad Rob Davis \quad  Mihaela van der Schaar \\
  University of Cambridge\\
  \texttt{bv292@cam.ac.uk} 
}

\begin{document}

\maketitle
\begin{abstract}
Tabular data is one of the most ubiquitous modalities, yet the literature on tabular generative foundation models is lagging far behind its text and vision counterparts. Creating such a model is hard, due to the heterogeneous feature spaces of different tabular datasets, tabular metadata (e.g. dataset description and feature headers), and tables lacking prior knowledge (e.g. feature order). In this work we propose LaTable: a novel tabular diffusion model that addresses these challenges and can be trained across different datasets. Through extensive experiments we find that LaTable outperforms baselines on in-distribution generation, and that finetuning LaTable can generate out-of-distribution datasets better with fewer samples. On the other hand, we explore the poor zero-shot performance of LaTable, and what it may teach us about building generative tabular foundation models with better zero- and few-shot generation capabilities.
\end{abstract}

\section{Introduction}
\parag{Motivation.} Foundation models (FMs) in the image and text domains epitomize the value of large scale training, achieving impressive results and pushing the boundaries of AI capabilities. In this work, we propose a tabular diffusion model that can be trained across vastly different datasets, and could possibly be the basis for a tabular foundation model.

Tabular foundation models, or Large Tabular Models (LTMs), lag far behind their text and vision counterparts, and existing LTM research focuses on representation and supervised learning \citep{van_breugel_why_2024}. Yet, despite this \emph{underrepresentation} in research, tabular data is \emph{ubiquitous} in the real world---including in medicine \citep{fatima_survey_2017,borisov_deep_2022}, finance \citep{dastile_statistical_2020}, census data \citep{office_for_national_statistics_2021_2021}, data science competitions \citep{kaggle_2017_2017, huang_tabtransformer_2020}, and natural science \citep{shwartz-ziv_tabular_2022}. A generative LTM could transform these fields, through enabling few- and zero-shot generation of synthetic data (e.g. for privacy, debiasing, data democratization, augmentation, adaptation, simulation, etc. \citep{van_breugel_beyond_2023}), better imputation, and providing a base model that can be finetuned to other tasks (e.g. representation learning or prediction). Few-shot capabilities are especially interesting for the tabular domain, where datasets are often smaller. In Box \ref{box:applications} we expand on applications of a generative LTM.

\parag{Aim.} Developing a generative LTM for tabular data is hard; different datasets cover vastly different features, there is no prior knowledge in the form of feature order, and data is messy---e.g. different datasets use different formats, contain different features, and can contain categorical, numerical, and missing values. As a result of this, training models meaningfully on large amounts of heterogeneous data remains underexplored.

We aim to take a step towards Large Tabular Models by developing a versatile tabular generative model with the following desiderata \citep{van_breugel_why_2024}:
\begin{enumerate}[label=\textbf{D\arabic*}]
    \item \label{d:crossdataset}\textbf{Cross-dataset generation.} We need a generative model to be able to generate different tables, which requires it being able to generate different features and variable number of features.
    \item \label{d:mixed}\textbf{Categorical and numerical feature generation.} Tables consist of combinations of both continuous and discrete data, and we want to be able to model each.\footnote{In contrast, an LLM-based approach like \citep{borisov_language_2023} implicitly models each numerical feature as a series of discrete tokens, see Section \ref{sec:related}.}
    \item \label{d:context}\textbf{Use textual context.} To achieve generating across tables, the model will need to understand contextual metadata---including the dataset description, feature names, and allowed categories (for categorical features).
    \item \label{d:equivariance}\textbf{Equivariance w.r.t. column order.} Table column order is usually arbitrary, so we want an LTM to be equivariant to this. In other words, define generative model $G:\mathcal{S}^L\times\mathcal{M}\rightarrow \mathcal{X}^L$ with $\mathcal{S}$ the column-wise input space (e.g. feature names), $\mathcal{X}$ the output space, and $\mathcal{M}$ metadata (e.g. dataset description); for input $\mathbf{s}\in \mathcal{S}^L$, $r\in\mathcal{M}$, and $T$ some permutation, we desire $G(T(\mathbf{s}),r)=T(G(\mathbf{s},r))$.
\end{enumerate}

\parag{Contributions.}
Our contributions are as follows:
\begin{enumerate}
    \item We introduce \name, a novel tabular diffusion model that satisfies these desiderata and can thus be trained across vastly different tabular datasets (including different features, number of features, and different types) and with context of table metadata. 
    \item We empirically show how \name outperforms existing generative models significantly on in-distribution generation, and when finetuned on new, out-of-distribution datasets.
    \item We explore the limitations of \name in terms of zero-shot capabilities. We argue that this stems from the limited pretraining of \name and discuss the challenges ahead for generative LTM research. Our findings points towards the central role that the training data should play in future research.
\end{enumerate}

\section{Related work} \label{sec:related}
\parag{Tabular single-dataset generative models.} Tabular data is challenging due to its lack of structural meaning (cf. images), mixed-type variables, and often limited size. A growing body of work is developing generative models tailored to tabular data \citep{choi_generating_2017,xu_modeling_2019,watson_adversarial_2023}, including diffusion/score-based models \citep{dieleman_continuous_2022,kotelnikov_hse_tabddpm_2023, zhang_mixed-type_2024}. Nonetheless, most methods ignore or naively one-hot-encode categorical data (e.g. feature names and categories), which loses information and context that could help overcome data availability issues (\ref{d:context}).\footnote{Of course, without cross-table training, it would not be possible to learn much from the dataset description, feature names, and categories, due to lack of diversity---each dataset provides just one description, and not many unique features or categories to learn from.} Naturally, methods developed for single datasets are also not designed for cross-datasets generation (\ref{d:crossdataset}, or equivariance w.r.t. column order (\ref{d:equivariance}).

\parag{LM-based approaches.}
Pretrained language models (LMs) may provide a solution to cross-dataset generation with (\ref{d:crossdataset}-\ref{d:equivariance}), as they contain general knowledge and can represent strings as numerical vectors in a space structured by language supervision. 
These works all convert the rows of tables into strings that can be processed by the LM. The string generated by the LM is then converted back to a tabular format. \citet{borisov_language_2023} finetune GPT-2 \citep{radford_language_2019} on tables. They aim to achieve approximate equivariance (\ref{d:equivariance}) through randomly permuting feature permutations, and sample tokens autoregressively from GPT's predicted token probabilities to model distributions (cf. sample only the most likely sentences). \citet{solatorio_realtabformer_2023} note that \citep{borisov_language_2023}'s approach retains GPT-2's original vocabulary, yet that most tokens may not appear in the tables of interest. They instead use a fixed-set vocabulary\citep{padhi_tabular_2021}, which reduces the chances of generating invalid samples and thus improves efficiency. Last, \citet{zhao_tabula_2023} show that an untrained, smaller LM can generate more accurate data and do so more cheaply than \citep{borisov_language_2023}. We note that in contrast to us, previous works \citep{borisov_language_2023,solatorio_realtabformer_2023,zhao_tabula_2023} train/finetune their model on just a single tabular dataset---i.e. they do not attempt cross-dataset training and do not aim to create a generative model that generalizes beyond the training data.

The advantage of LM-based generation is the simplicity and the fact that it does not require any manual preprocessing of data, since the raw data can be presented to the LM directly (e.g. prompting and in-context examples). However, adapting LMs as LTMs comes with serious disadvantages \citep{van_breugel_why_2024}. First, LMs are expensive, during both finetuning and inference. A 1-min training job for CTGAN, takes a whopping $540\times$ longer for \citep{borisov_language_2023} (see their Appendix B.5, Table 6). 
One of the core sources of this inefficiency is the linearization of table rows as sentences, and generating these autoregressively. For tabular data, where many columns are typically numerical, this is problematic---it means that a single numerical variable is implicitly modelled as an autoregressive series of categorical variables (e.g. $1.23$ is modelled as $1\rightarrow .\rightarrow 23$). Hence, generating this number with an LM requires 3 expensive forward calls to the model (plus one for the field separation token that follows). Fitting the whole data in context might also become difficult with datasets containing hundreds of columns. Furthermore, the multinomial training objective of LM is not apt at approximating continuous distributions---e.g. \citet{renda_can_2023} show how LMs do not accurately generate simple distributions (e.g. uniform) and \citet{van_breugel_why_2024} visualize how modelling a simple Gaussian distribution with an LM is non-trivial. By opting for an end-to-end numerical representation of the data and a diffusion model framework, \name circumvents these limitations.

\section{Method} \label{sec:method}
\parag{Summary.} We propose \name, which satisfies the desiderata. \name uses an encoder-only transformer as backbone for mixed-type diffusion. The input of the model consists of noised feature values (i.e. from the forward noising process), but also the dataset description, column names, conditioning mask (in case of conditional generation), and boolean missingness mask. All metadata strings (e.g. dataset description and feature names) and categorical features are first encoded using a pretrained LLM into a fixed-length embedding. All inputs are subsequently mapped to a common hidden space using separate element-wise MLPs, and added to provide the direct input of the transformer. The output of the transformer is decoded on a feature-by-feature basis, where categorical features are mapped to probabilities over the original features. See Figure \ref{fig:architecture}.

\subsection{Set-up and notation}
We index datasets with $k$, samples with $i$, and features with $j$.\footnote{To avoid excessive clutter, we will leave out indices when they are irrelevant or clear from context.} Let us assume we have access to a metadataset $\mathcal{D} = \{D\sk\}_{k=1}^{n_D}$, where each dataset $D\sk$ consists of a description, column names, and the data itself. Let $d_k\in \mathbb{N}$ denote the number of columns for dataset $k$. Feature spaces will generally differ for each dataset, hence let us define the $j$th feature space for dataset $k$ as $\mathcal{X}\sk_j$---i.e. for categorical data we let this be a finite set of strings, for numerical columns it may be some subset of $\bbR$. Let us denote the set of all strings with $\mathcal{S}$. We can thus write, $D\sk = (r\sk, \mathbf{s}\sk, \mathbf{X}\sk)$, with $r\sk\in\mathcal{S}$ the dataset description, $\mathbf{s}\sk\in \mathcal{S}^{d_k}$ feature names, and samples $\mathbf{X}\sk_i\in \prod_{j=1}^{d_k} \mathcal{X}\sk_j$. Finally, let us denote the indices of the categorical features by $\catindices$.

\subsection{Generative model choice}
We use a diffusion model \citep{ho_denoising_2020} (i.e., score-based model \citep{song_generative_2019}) for modelling the data distribution. Diffusion models have increased in popularity thanks to their capability to produce samples of higher quality than GANs and VAEs. We use the discrete-time formulation with Denoising Diffusion Implicit Model (DDIM) noise scheduler \citep{song_denoising_2021}.

To satisfy the desiderata, however, we need to think carefully about what architecture backbone we use for the diffusion model, how we model both continuous and categorical variables, and how the tabular metadata fits in. 
Let us look into each element of \name in more detail, and how it helps to satisfy desiderata \ref{d:crossdataset}-\ref{d:equivariance}.

\subsection{Satisfying desiderata}
\subsubsection{Transformer backbone (\ref{d:crossdataset}, \ref{d:equivariance})} 
We use an encoder-only transformer at the core of \name. The transformer backbone is useful for our setting, as it allows variable-length input and output---this allows us to train across datasets (\ref{d:crossdataset}). We \emph{do not} use a positional encoding for the input, such that the transformer is equivariant w.r.t. the input (\ref{d:equivariance}). In contrast, a recurrent architecture would not necessarily satisfy this equivariance property.\footnote{Importantly, removing the positional encoding means the transformer cannot know which column is which. We remove this undesirable symmetry by adding column information to each input element, in the form of a feature name embedding---see Section \ref{sec:context}.} We denote the transformer's hidden dimension by $d_h$.

\subsubsection{Using context (\ref{d:context})} \label{sec:context}
To make use of context (\ref{d:context}), we will encode the category names, column names, and dataset description using a pretrained LLM encoder. Note that the dataset description is the same across all features, but that the other metadata are defined on the feature level. By using a pretrained LLM encoder, we avoid having to learn embeddings for descriptions and categories from scratch. It also enables us to use textual similarities, even if our tabular data is limited. For example, one dataset may contain column ``gender'' and another ``sex'', and we need not learn from the data alone that these are closely related.

We denote the pretrained LLM by $\Enc$. Text encodings are fixed length $d_{f}$ for any input string $\mathcal{S}$, such that $\Enc:\mathcal{S}\rightarrow \bbR^{d_{f}}$. Whenever we encode text data, we will denote this with a bar, e.g. $\bar{\mathbf{r}}\sk$, as follows:
\begin{enumerate}
    \item Each string dataset description $r\sk$ is encoded by the LLM into a vector $\bar{\mathbf{r}}\sk$.
    \item Each categorical variable is converted into a string ``[column name $s$] is [category $c$]". Categories $(c_1,...,c_n)$ are then encoded by the LLM and stacked into an \emph{embedding matrix} $\bar{\mathbf{C}}^{(k)}_j$
\begin{equation} 
\label{eq:embedding_matrix}
    \bar{\mathbf{C}}^{(k)}_j = (\Bar{\mathbf{c}}_1,...,\Bar{\mathbf{c}}_{n}).
\end{equation}
    \item Feature names $(s_j)_j$ are encoded by the LLM into vectors $(\bar{\mathbf{s}}_j)_j$
\end{enumerate}
We condition on the encoded metadata by mapping the dataset description $\bar{\mathbf{r}}$ and feature embeddings $\bar{\mathbf{s}}_j$ to size $d_h$ using shallow MLPs $g_r,g_s$ and adding it to the transformer input (see Figure \ref{fig:architecture} in blue).

For the encoder $\Enc$, we use model \texttt{WhereIsAI/UAE-Large-V1} for three reasons: it is open source and easily available,\footnote{\href{https://huggingface.co/WhereIsAI/UAE-Large-V1}{https://huggingface.co/WhereIsAI/UAE-Large-V1}}; it achieved SOTA on the Massive Text Embedding Benchmark (MTEB) \citep{muennighoff_mteb_2023} until December 31 2023; and it is very lightweight (size: 1.34 GB). All encodings are cached to disk before training and testing to avoid LLM forward calls during training.

\subsubsection{Mixed-type variables (\ref{d:mixed})}
In contrast to standard LLM transformers and diffusion models, we need to generate numerical values as well as categoricals (\ref{d:mixed}). Let us start with the numerical variables.

\parag{Numerical variables.} For numerical variables we can use relatively standard diffusion. Given some time step $t\in T$, the forward denoising process is applied to all numerical features $x_j$, giving the noisy $(x_j)_t$. A shallow MLP $g^i_n: \bbR\rightarrow \bbR^{d_h}$ is used to map each numerical variable to the transformers input space. The transformer is subsequently applied to the sequence of all variables. Subsequently, each sequence element in the transformer's output corresponding to a numerical variable is mapped back to scalars using output network $g^o_n:\bbR^{d_h}\rightarrow \bbR$. For each numerical $(x_j)_0$, this yields output $(\hat{x}_j)_0$, and the loss (e.g. sample-loss, $\epsilon$-loss, or $v$-prediction) is computed per feature.

\parag{Categorical variables.} To handle categorical variables, we draw inspiration from Continuous Diffusion for Categorical Data \citep{dieleman_continuous_2022}. The original CDCD learns high-dimensional embeddings for each category separately, before applying $L_2$ normalization to each embedding (after which noise is added for the denoising). During training, CDCD adds noise to these categorical embeddings (identical to continuous diffusion), and trains a model that predicts probabilities for the original categories. This model is trained using some classification loss, e.g. cross-entropy. During sampling, the outputted probabilities are used to predict the score in the continuous space. Specifically, they use \emph{score interpolation}---which is effectively a weighted mean of all category embeddings, where weights are given by the probabilities predicted by the score model.

In our case, we do \emph{not} want to learn embeddings and predict probabilities for all possible categories in all datasets. By learning an embedding separately for each category, this loses context that is captured in the category name (e.g. how categories \textit{female} and \textit{woman} are related). It also makes it hard to scale to very large number of total categories, especially if some of these categories are hardly ever observed. Instead, we use the pretrained LLM $\Enc$ for acquiring ${d_f}$-dimensional embeddings for each category $c$, which \emph{do} carry the contextual knowledge of the category name.\footnote{As discussed in Section \ref{sec:context}, for all categorical features we embed the categorical $c$ as $\Bar{\mathbf{c}} = $``[feature name] is [$c$]".} We fine-tune these embeddings using a shallow MLP followed by an $L_2$ normalization layer (similar to CDCD), which we together denote by $g_f:\bbR^{d_f}\rightarrow S^{d_f}$.\footnote{Without the normalization layer, it would be difficult for the model to give embedded categories with small $L_2$-norm a high probability (see Eq. \ref{eq:attent_prob}). We add the shallow MLP's finetuning, to avoid projecting embeddings that are quite different in $\bbR^d$ to the same point on the unit sphere.} The forward diffusion process corrupts these vectors by adding Gaussian noise, giving $\Bar{\mathbf{c}}_t$. Similar to the numericals, we use a network $g^i_c:\bbR^{d_f}\rightarrow \bbR^{d_h}$ to map these noisy embeddings to the transformer's input space, and network $g^o_c:\bbR^{d_h}\rightarrow \bbR^{d_f}$ is applied to the categorical sequence elements in the transformer's output---giving predicted embeddings in $\bbR^{d_f}$. 

One key question remains: considering each categorical feature may have a different number of categories, how do we map the transformer output (in $\bbR^{d_f}$) back to the individual categories? We achieve this using an attention-like layer. Letting $\hat{\mathbf{c}}_j\in \bbR^{d_f}$ be the output of the model, we let the final predicted probability for some category $i$ be proportional to the similarity of the generated vector and the original categories' embeddings (see Eq. \ref{eq:embedding_matrix}):
\begin{equation}
\label{eq:attent_prob}
    % (x_0)^k_j=c|\mathbf{x}_t
    p(c_i) = \text{softmax} (g_f(\bar{\mathbf{C}})^T \hat{\mathbf{c}})_i, \forall i,
\end{equation}
thereby allowing us to predict probabilities for each datasets' and categoricals' categories individually.
Identical to CDCD, we use cross-entropy loss $\mathrm{CE}$ for the categorical variables and use score interpolation. Combining the numerical and categorical losses, we get:
\begin{equation}
    \mathcal{L}(x, \hat{x}) = \frac{1}{d_k}\left[\sum_{j\in \catindices} \mathrm{CE}(x_j, \hat{x}_j)+\sum_{j\not\in\catindices} \mathrm{MSE}(x_j, \hat{x}_j)\right]. 
\end{equation}

\begin{figure*}
    \centering
    \includegraphics[width=0.8\textwidth]{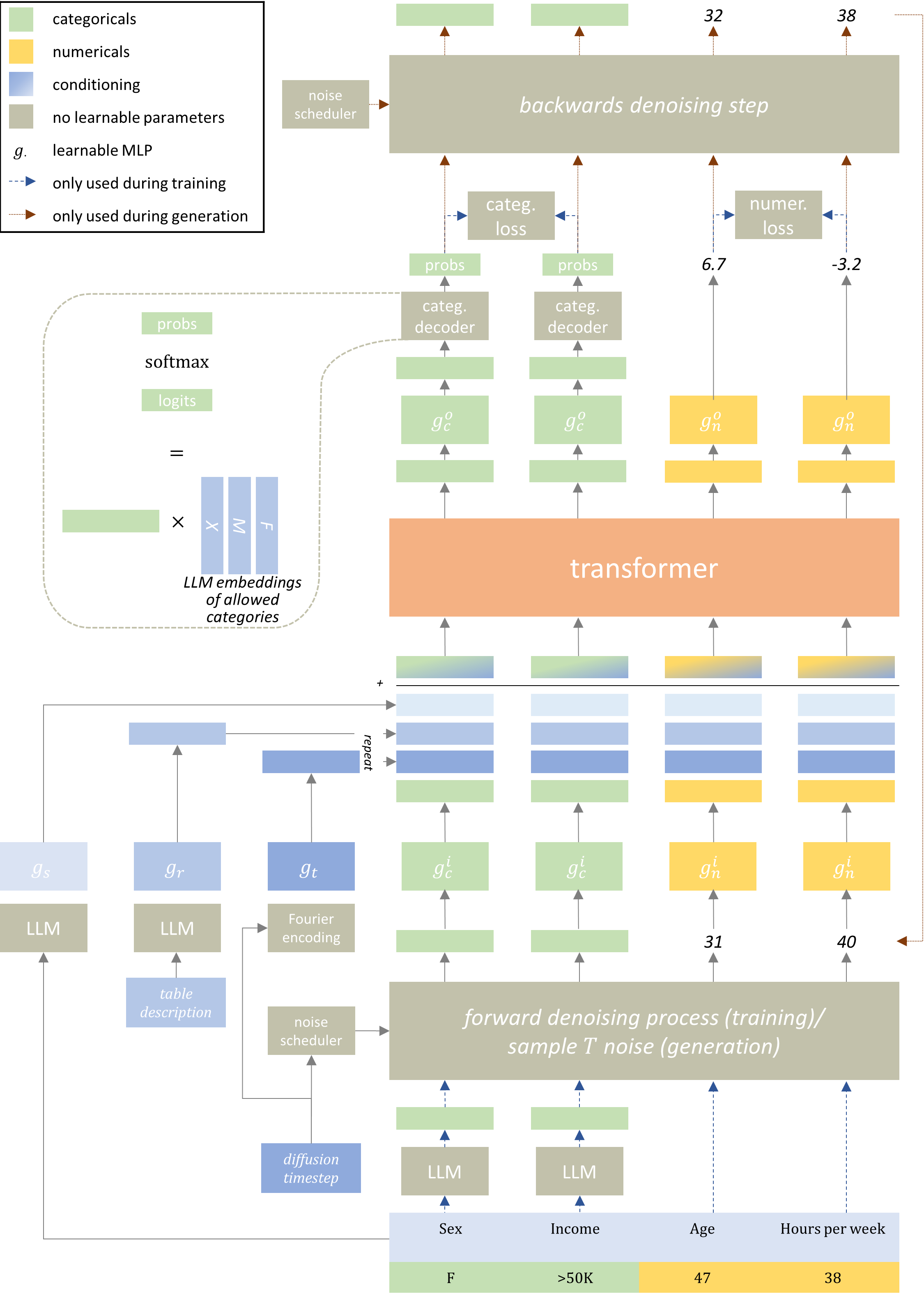}
    \caption{The diffusion pipeline, for both training and generation. Note that the whole pipeline is flexible with respect to the number of numerical and categorical features as input. The LLM is encoder is frozen and the transformer is an encoder-only model without positional encodings or causal masking. Additional conditioning (e.g. missingness mask, conditioning information) can be trivially added to the transformer input, or through cross-attention layers. 
    % See Alg. \ref{alg:pseudo} for pseudo code.
    }
    \label{fig:architecture}
\end{figure*}

\parag{Bringing it together.} Overall, the model thus consists of a number of independent MLPs $g_\cdot$, that allow us to bring differently sized variables (dataset description, feature column name embeddings, categorical features, numerical features) all to the same space $d_h$. The input of the transformer consists of a sequence of inputs $h_j\in\mathbb{R}^{d_h}$, $j=1, ..., d_k$ , with $h_j = g^i_c((\Bar{x}_j)_t) + g_r(\Bar{r}) + g_s(\Bar{s}_j) + g_t(t)$ for categoricals, and $h_j = g^i_n(x_j)_t) + g_r(\Bar{r}) + g_s(\Bar{s}_j)+ g_t(t)$ for numericals. The transformers output sequence is again decoded using independent shallow networks $g_c^o, g_n^o$, to provide score estimates for both numerical and categorical variables. See Figure \ref{fig:architecture} for a full overview.
% Algorithm \ref{alg:pseudo} for the full training pseudocode.

\section{Experiments}
In this section we explore the tabular data generation capabilities of \name. We show how \name achieves significantly better in-distribution generation, as well as few-shot out-of-distribution generation through minor finetuning. These observations suggest that \name's data generation capabilities generalize across tables. At the frontier of these capabilities, we discuss zero-shot tabular data generation as a limitation of \name. We explore how this relates to the limited diversity and quality of the training data, drawing lessons for future research into generative LTMs.

\textbf{Data.} We combine curated metadatasets \citep{bischl_openml_2021} and \citep{fischer_openml-ctr23_2023} from OpenML, which respectively cover classification  and regression tasks. After removing image datasets, these total 83 sets (678k samples, 3021 unique columns)---see Figure \ref{fig:data} for number of samples and features per dataset. We split up these datasets into in-distribution $\mathcal{D}_{\mathrm{id}}$ and out-of-distribution $\mathcal{D}_{\mathrm{ood}}$, including respectively 78 and 5 datasets. In contrast to other large metadatasets (e.g. TabLib \citep{eggert_tablib_2023}) that may contain (processed) duplicate sets, the datasets we use are curated, hence there is no data leakage. All experimental details can be found in Appendix \ref{app:details}.

\subsection{In-distribution generation}
We use independent train-val-test splits of each set in $\mathcal{D}_{\mathrm{id}}$, and train \name on all training sets in $\mathcal{D}_{\mathrm{id}}$. We compare the generation quality of \name to popular synthetic data baselines. We choose tabular generators in different classes, including TVAE, CTGAN \citep{xu_modeling_2019}, TabDDPM (diffusion) \citep{kotelnikov_hse_tabddpm_2023}, and ARF (tree-based) \citep{watson_adversarial_2023}. Note that for each of these models, we train a new model on each individual dataset and for each seed. We compare downstream model train-on-synthetic-test-on-real AUC performance (of an XGBoost classifier and taking the label of the original dataset); synthetic data metrics precision and recall \citep{sajjadi_assessing_2018,kynkaanniemi_improved_2019}; and density and coverage \citep{naeem_reliable_2020}. Note that precision and density aim to quantify realism or fidelity of samples, whereas recall and coverage measure diversity.

As we observe in Table \ref{tab:overall_iid}, \name achieves significantly better results than baselines on most metrics. The closest competitor is the tree-based ARF, which scores comparably on downstream AUC and recall but significantly poorer on the other metrics. Metric \textit{Recall} gives a high score when data samples are very diverse (i.e. the synthetic distribution has a large support), but this does necessarily reflect well on data quality (see Fig. 1b in \citep{naeem_reliable_2020}). This might explain why recall is high for all methods, even though they fail on other metrics. 
\begin{table}[hbt]
    \centering
\caption{Average performance of each generative model across 78 OpenML datasets in $\mathcal{D}_\mathrm{id}$. We report the mean and standard deviation over 5 random seeds. Downstream AUC is computed for an XGBoost model trained on synthetic, tested on real data (only including classification datasets). \name is significantly better on all metrics except recall.}
    \label{tab:overall_iid}
\begin{tabular}{llllll}
\toprule
\emph{Generator}  & \emph{Down. AUC} [$\uparrow$]&       \emph{Density} [$\uparrow$] &      \emph{Coverage} [$\uparrow$] &     \emph{Precision} [$\uparrow$] &        \emph{Recall} [$\uparrow$] \\
\midrule
     Real &  0.895 (0.007) & 0.961 (0.005) & 0.955 (0.002) & 0.953 (0.003) & 0.945 (0.001) \\ \hline 
      ARF &  0.853 (0.009) & 0.700 (0.008) & 0.832 (0.003) & 0.836 (0.002) & \textbf{0.942 (0.001)} \\
    CTGAN &  0.842 (0.008) & 0.584 (0.006) & 0.741 (0.003) & 0.720 (0.002) & 0.912 (0.004) \\
     TVAE &  0.828 (0.007) & 0.739 (0.010) & 0.805 (0.004) & 0.812 (0.007) & 0.877 (0.002) \\
  TabDDPM &  0.757 (0.014) & 0.414 (0.015) & 0.535 (0.022) & 0.469 (0.021) & 0.823 (0.011) \\
  \textbf{\name} &  \textbf{0.874 (0.006)} & \textbf{0.865 (0.009)} & \textbf{0.900 (0.001)} & \textbf{0.866 (0.004)} & 0.900 (0.002) \\
\bottomrule
\end{tabular}
\end{table}

\begin{figure}[hbt]
    \centering
    \begin{subfigure}[b]{0.5\textwidth}
    \includegraphics[width=\textwidth]{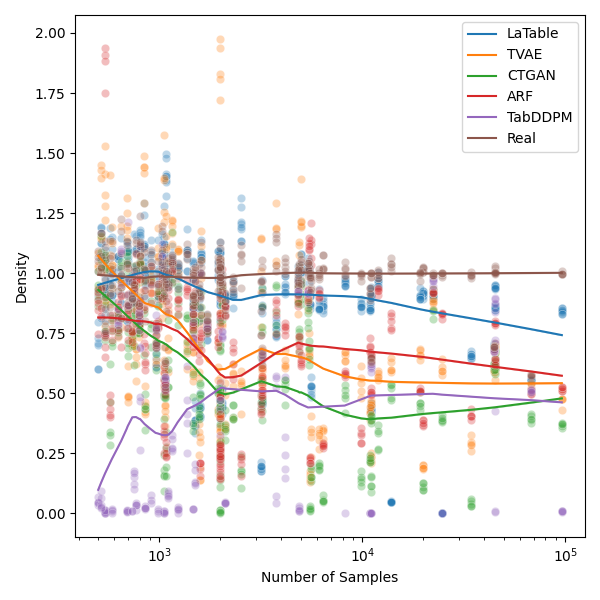}
    \caption{Density}
    \end{subfigure}~
    \begin{subfigure}[b]{0.5\textwidth}
    \includegraphics[width=\textwidth]{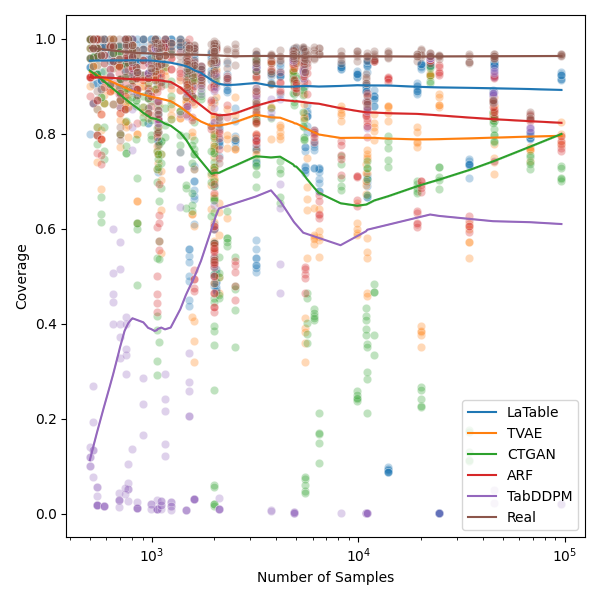}
    \caption{Coverage}
    \end{subfigure}
    \caption{\textbf{\name outperforms baselines especially for smaller datasets.} Generation results for different generative models across 78 datasets (sorted by size), with 5 seeds per (model, dataset) and a LOWESS curve fitted for smoothing.}
    \label{fig:iid_n_samples}
\end{figure}

It is insightful to more closely analyze the origin of \name's gains against the baselines. In Figure \ref{fig:iid_n_samples} we plot the metrics for each individual dataset and as a function of the number of samples in each dataset. We observe that \name performs more reliably for datasets with smaller sample size, and on average better across the board. This can be understood as a consequence of \name being trained on other datasets besides the small datasets. Hence, this suggests that \name outperforms the baselines by transferring data generation capabilities across tables. We note that despite hyperparameter tuning attempts, TabDDPM performed poorly for small datasets.

\subsection{Zero- and few-shot out-of-domain generation} \label{sec:ood}

An important characteristic of foundation models is their ability to be adapted to new datasets, preferably using relatively little data (i.e. few-shot). In this section we thus aim to use the trained model $G$ on some target datasets $D_T \in \mathcal{D}_{\mathrm{ood}}$ that it has not seen during training, and see how it compares to baselines.

For \name, we would ideally want good zero-shot performance---i.e. to generate new data based on the new data's feature names and metadata alone. However, we will show that this constitutes the limits of \name's capabilities and pretraining. A more realistic scenario is to adapt \name to the new datasets through finetuning. To measure the effect of how much data is needed for finetuning, we artificially reduce the target dataset to a smaller number of samples. We compare to baselines that are trained on the target dataset alone.

\begin{figure}[hbt]
    \centering
    \begin{subfigure}[b]{0.5\textwidth}
    \includegraphics[width=\textwidth]{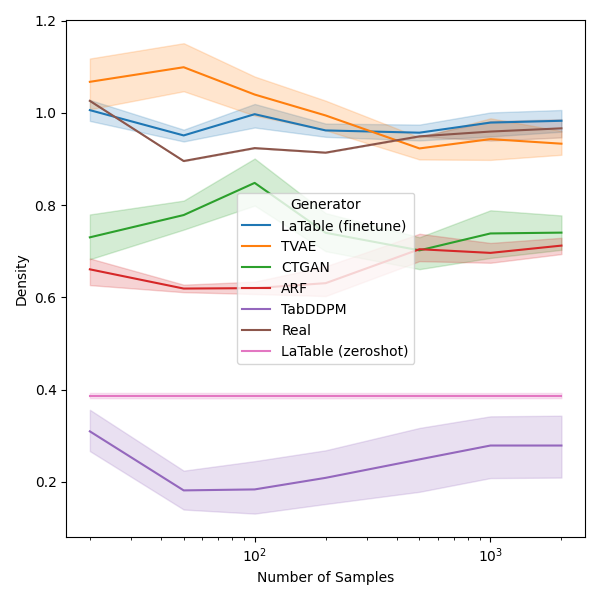}
    \caption{Density}
    \end{subfigure}~
    \begin{subfigure}[b]{0.5\textwidth}
    \includegraphics[width=\textwidth]{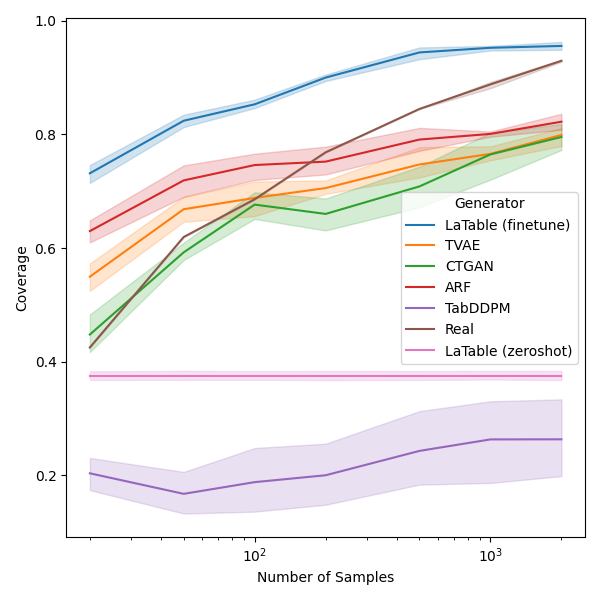}
    \caption{Coverage}
    \end{subfigure}
    \caption{\textbf{A finetuned \name outperforms baselines significantly on out-of-distribution datasets with few samples.} Generation performance on out-of-distribution datasets as a function of the training dataset size. Metrics are averaged over 5 datasets in $\mathcal{D}_{\mathrm{ood}}$, where datasets are reduced to $n_{samples}$. Mean and error bars (95\%) are computed over 5 seeds.}
    \label{fig:ood_n_samples}
\end{figure}

In Figure \ref{fig:ood_n_samples} we plot the performance of \name and baselines as a function of the number of target samples, which is averaged over 5 seeds and 5 different datasets in $\mathcal{D}_{\mathrm{ood}}$. Baseline \emph{Real} is again the real training data, but in contrast to last section, this dataset is now very small. 

As aforementioned, the zero-shot performances of \name does not match the baselines---it has not seen enough data during training to generalize well. When finetuned, however, \name exhibits a more promising behavior. On \emph{Density} (i.e. fidelity) it achieves an almost perfect score of 1, on par with real data. Among the baselines,  TVAE stands out. However, its density larger than 1 suggests that the model collapsed to a high-density region of the distribution. When we look at \emph{Coverage} (i.e. diversity), \name appears even stronger. We hypothesize that due to the pretraining and its metadata, it is able to generate more diverse realistic samples in a few-shot setting. This hypothesis is reinforced by noting that \name achieves significantly better diversity than the real data---this is possible, considering the real training data includes just a few samples.

\section{Challenges to generalization} \label{sec:scaling}
Let us explore possible directions for future research into better few- and zero-shot \name.

\parag{Why \name achieves poor zero-shot performance.}
We cannot expect good zero-shot performance of \name, as $\mathcal{D}_{\mathrm{ood}}$ is not covered by $\mathcal{D}_\mathrm{{id}}$. In Appendix \ref{app:comparing_features} we analyse the features of $\mathcal{D}_\mathrm{id}$ and $\mathcal{D}_\mathrm{ood}$. We observe that most features in $\mathcal{D}_\mathrm{ood}$ do not have a close feature in $\mathcal{D}_\mathrm{id}$. Similarly, even for the features that are observed during training, there may be a significant shift in the feature distribution---let alone, how it interacts (e.g. correlates) with other features.

\parag{Improving out-of-distribution performance through scaling.}
The straightforward solution to this problem is to increase the training dataset size---this has been one of the key drivers behind text and vision FM progress \cite{brown_language_2020,kaplan_scaling_2020, rombach_high-resolution_2022, hoffmann_empirical_2022}. Multiple large tabular metadatasets exist: WebTables \citep{lehmberg_large_2016} consists of 10M crawled HTML tables, WikiTables \citep{bhagavatula_tabel_2015} includes 580K Wikipedia tables, and the recent Tablib \citep{eggert_tablib_2023} includes 627M crawled tables with 827B context tokens. 

Though these metadatasets are large, unfortunately they primarily include a significantly different type of table. In Appendix \ref{app:other_sources} we show they consist of smaller tables, both in terms of features and samples (see Figure \ref{fig:compare_sources}). We also show this affects \name's generation. WikiTables is curated, covers a wide range of topics with context from the Wikipedia article, and is guaranteed to not have overlap with $\mathcal{D}_\mathrm{ood}$ (vs Tablib or WebTables, which may bias our results), hence let us consider this for training \name. We discard any table with less than 20 samples, which yields 100k+ tables, and repeat the experiment from Section \ref{sec:ood} for a model trained on WikiTables. Though we observe (Figure \ref{fig:wikitables}) comparable performance to the original \name when finetuned, the zero-shot performance is significantly worse.

\parag{Going forward.} Our findings emphasize the importance of good quality data in improving generalization performance of \name. The issue of data quality may be resolved by better understanding of large metadatasets like TabLib, and using heavy-handed curation and selection. To avoid data leakage between the training and evaluation sets, this also requires robust tools for identifying and removing duplicates, which is hard for tabular datasets since preprocessed and filtered copies are common. Another possible avenue for future work is to use more specialized databases for training, e.g. of high-throughput biological data \cite{clough_gene_2016}.

\section{Discussion} \label{sec:discussion}
\parag{Limitations.} 
\name can be extended and improved in multiple ways:
\begin{enumerate}
    \item \textbf{Scale and data.} As discussed in the previous section, \name is a relatively small model trained on a relatively small dataset. Scaling up is one of the key challenges for future research.
    \item \textbf{Extending variable and table types.} We restrict ourselves to numerical and categorical data, but future work could consider datetime, full string descriptions, time-series, and relational databases to increase applicability and impact.
    \item \textbf{Bias.} In this work, we did not explore possible bias in the data and pretrained LLM, or how \name copies it. More research into bias is required before applying \name to the real-world, see broader impact below.
\end{enumerate}

\parag{Broader impact.} We believe the broader impact of \name, and continued work in this area, is primarily positive. \name can enable better synthetic data using fewer samples, which could promote more responsible AI---e.g. to improve \textit{minority representation using data augmentation} \citep{chawla_smote_2002, das_conditional_2021, van_breugel_can_2023}, \textit{ML model testing through data simulation} \citep{van_breugel_can_2023}, and \textit{data democratization through private synthetic data} \citep{ho_dp-gan_2021}. The latter is especially interesting, since few-shot generation may require a smaller privacy budget than standard synthetic data---after all the model need not learn the full distribution from the private data and may be less likely to memorize this data \citep{van_breugel_membership_2023}.

Nonetheless, we need to acknowledge the risk: \name may make errors and may reflect or exacerbate societal biases that are present in the data, or in the pretrained LLM embeddings. More research into possible biases is required. The quality and fairness of generated data should always be evaluated before applying \name to real-world sensitive settings like healthcare and finance.

\bibliographystyle{unsrtnat}
\bibliography{references_zotero}

%%%%%%%%%%%%%%%%%%%%%%%%%%%%%%%%%%%%%%%%%%%%%%%%%%%%%%%%%%%%%%%%%%%%%%%%%%%%%%%
%%%%%%%%%%%%%%%%%%%%%%%%%%%%%%%%%%%%%%%%%%%%%%%%%%%%%%%%%%%%%%%%%%%%%%%%%%%%%%%
% APPENDIX
%%%%%%%%%%%%%%%%%%%%%%%%%%%%%%%%%%%%%%%%%%%%%%%%%%%%%%%%%%%%%%%%%%%%%%%%%%%%%%%
%%%%%%%%%%%%%%%%%%%%%%%%%%%%%%%%%%%%%%%%%%%%%%%%%%%%%%%%%%%%%%%%%%%%%%%%%%%%%%%
\newpage
\appendix

\section{Broader impact of generative LTMs}
The potential impact of a foundational tabular models is large. Generative models require a lot of data to be trained, inhibiting their use in smaller domains. On the other hand, if we would able to learn from multiple datasets, combine knowledge, and translate knowledge from one dataset's domain to another, this could enable generating data and making predictions with less data. We highlight some applications in Box \ref{box:applications}.

% \name has extensive applications for responsible AI:
% \begin{enumerate}
% \item \textbf{Inclusivity and Representation}: Through generating synthetic data we can improve underrepresentation in existing datasets, e.g. using data augmentation for generating marginalized groups few-shot. This in turn can allow better downstream science and ML development for these groups.
% \item \textbf{Robustness through Data Simulation}: Generated datasets may approximately simulate unseen or scarcely seen scenarios. This could help in model development, probing (e.g. for fail cases), and monitoring. 
% \item \textbf{Privacy and Wider Data Access}: By creating datasets that can mimic sensitive or restricted data without using the actual data, we can retain privacy while democratising data access.
% \item \textbf{Accelerate Discovery in Natural and Life Sciences}: Custom synthetic datasets open the doors to tailoring or repurposing data to scientific research, facilitating hypothesis generation and scientific discovery.
% \end{enumerate}

\begin{displaybox}{Applications of generative Large Tabular Model (LTM).}
\label{box:applications}
\small
\emphcol{(i)} Because LTMs have seen a lot of data already, their adaptation to new datasets through transfer learning is likely to be more data-efficient (e.g. it can leverage feature correlations that hold across datasets). \emphcol{(ii)} An LTM would enable to generate data for small datasets or datasets with underrepresented groups, by using the knowledge from other datasets. 
\emphcol{(iii)} Data augmentation using knowledge from other datasets could also allow more powerful science---implicitly leveraging data from other domains that are usually not considered.
\emphcol{(iv)} A conditional generative model (i.e. that allows generation based on a subset of the variables), allows (probabilistic) zero or few-shot prediction. These predictions can be more powerful than just training a predictive model (because the LTM may have seen data from different, but similar domains), and can give uncertainty estimates (because generative models are probabilistic).
\emphcol{(v)} Similarly, a conditional LTM permits (multiple) imputation of missing data\citep{rubin_multiple_1991}---few-shot.
\emphcol{(vi)} Because LTMs can be trained on public data only and require fewer data point to adapt to a new task, they can be used for private generation or prediction on new datasets while requiring a smaller privacy budget---e.g. no need to use private data to learn ``global'' correlations that can be learnt from other datasets. 
\end{displaybox}
% This would further promote data inclusivity, both in terms of marginalized group representation (by generating additional data for these groups) and data access (by adding privacy constraints to the generative model); enable more robust ML models through simulating unseen scenarios; and provide tailored or repurposed datasets for scientific discovery.

\section{Implementation details} \label{app:details}
\textbf{Encoding example.} As mentioned in \ref{sec:context}, we use text encoder \texttt{WhereIsAI/UAE-Large-V1} for encoding all string values. Let us consider an example:
\begin{table}[h!]
    \centering
    \caption{This describes the dataset.}
    \begin{tabular}{c|c|c}
         Age & Gender & Income \\ \midrule
         42 & Male & 30,000 \\
         33 & Female & 46,000 \\
    \end{tabular}
    \label{tab:example encoding}
\end{table}

In this case, we encode and cache ``This describes the dataset'' ( $\Bar{\mathbf{r}}$), feature names ``Age'', ``Gender'', and ``Income''  ($\bar{\mathbf{s}}_j$), and ``Gender is Male'' and ``Gender is Female''. 

\subsection{Data} \label{app:data_details}
\begin{figure}
    \centering
    \includegraphics[width=0.5\textwidth]{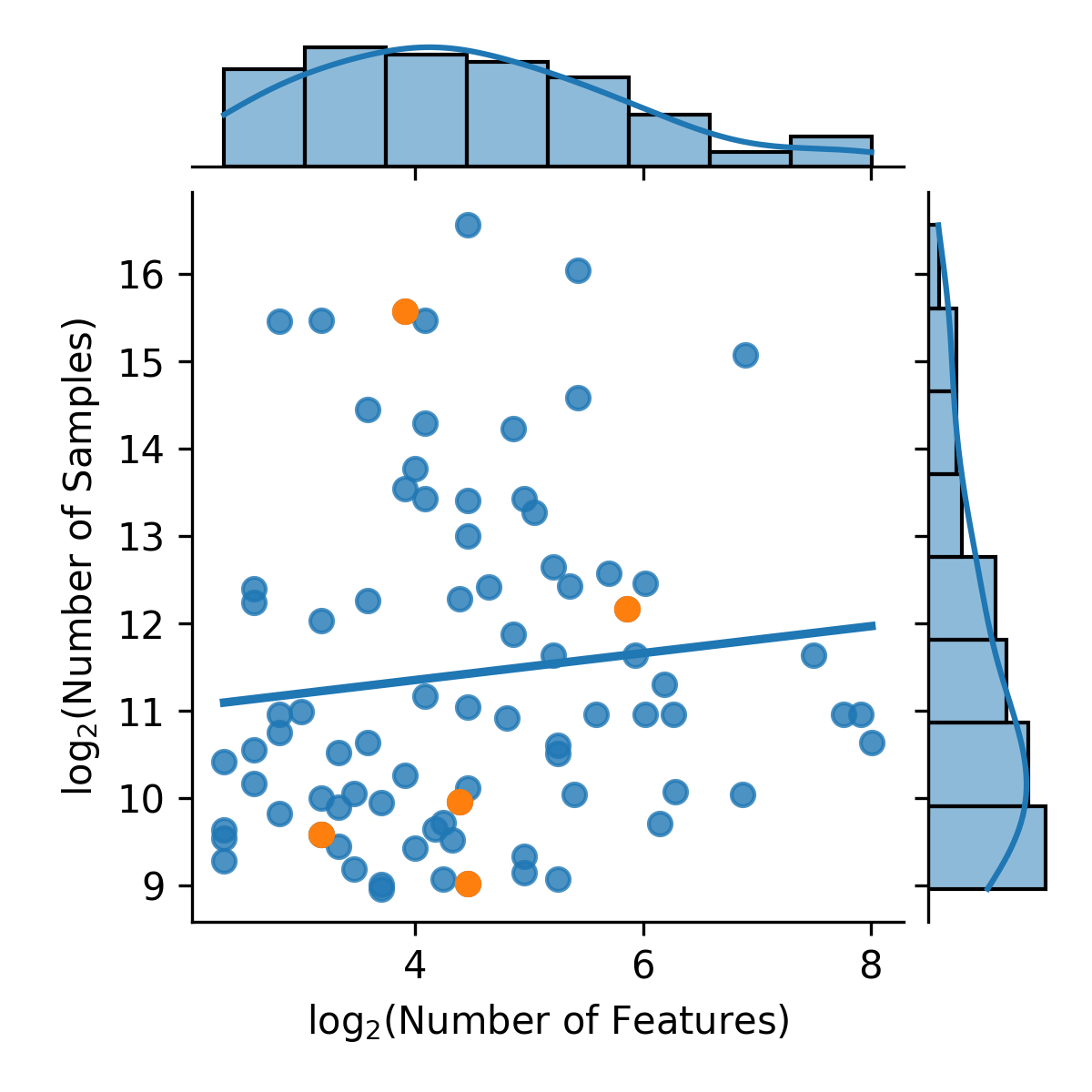}
    \caption{Number of samples per dataset plotted against number of features, with in-distribution datasets $\mathcal{D}_{\mathrm{id}}$ in blue and $\mathcal{D}_{\mathrm{ood}}$ in orange. \textit{Note: double log-scale}.}
    \label{fig:data}
\end{figure}

Data is loaded from OpenML dataset benchmarks \citep{fischer_openml-ctr23_2023, bischl_openml_2021}. Datasets with more than 500 columns are discarded as these correspond mostly to image datasets, which results in 83 datasets. We take 78 datasets for training \name ($\mathcal{D}_{\mathrm{id}}$) and 5 datasets for testing out-of-distribution performance ($\mathcal{D}_\mathrm{ood}$)---the latter includes OpenML datasets with dataset IDs 31 (credit-g), 37 (diabetes), 44 (spambase), 1063 (kc2), and 1570 (adult). Additional train-val-test splits are created for each dataset individually, and all training sets in $\mathcal{D}_{\mathrm{id}}$ are combined for training \name. The number of samples and features per dataset is visualized in Figure \ref{fig:data}. Numerical features are normalized.

\subsection{Model and training} \label{app:model_details}
For \name, we use the architecture described in Section \ref{sec:method} and visualized in Figure \ref{fig:architecture}. All MLPs $g_\cdot$ consist of 3 layers, and the transformer consists of 10 layers. All hidden layers are of dimension $1024$. We train and finetune for 500 epochs using Adam, with a cosine scheduler and learning rate $5\cdot 10^{-5}$. These values were chosen based on the in-distribution validation loss and validation generation performance, and are kept the same across all experiments. During training we weight points by their respective dataset size to ensure small datasets are not ignored, with dataset $k$'s weight $w^{(k)}\propto \frac{1}{\sqrt{|D^{(k)}|}}$. We use $T=200$ inference steps for DDIM, and use the standard Diffusers implementation of DDIM. Note that zero-shot generation, we \emph{do} assume we know the mean and standard deviations of numerical variables (as we retain scaling) and we \emph{do} assume which categories are allowed to be generated for each categorical feature (i.e. $\mathbf{\hat{C}}$ is provided).

\subsection{Baselines and metrics.} We use the Synthcity library \citep{qian_synthcity_2023} for baselines and metrics. Baselines use standard recommended settings for all datasets. For TabDDPM, we use 1000 denoising steps. We have tried changing the MLP size, but this did not yield reliably better results. To compute the metrics, we generate $\min \{|D_\mathrm{test}|,5000\}$ samples. For the ``Real'' baseline in Figure \ref{fig:ood_n_samples}, we oversample the small real datasets to acquire the required number of points for computing the metrics.

\subsection{Compute} \label{app:compute_details}
The model as described above can be trained on a 16GB GPU with batch size 32. Training \name on all datasets in $\mathcal{D}_\mathrm{id}$ takes around 2 days on a single GPU. Finetuning takes about 3 seconds per sample (e.g. 5 minutes for 100 samples). Generation speed is around 30 samples per second. For model development, we trained 33 models with different hyperparameters on all of $\mathcal{D}_{id}$.

\section{Improving out-of-distribution generation (accompanying Section \ref{sec:scaling})} \label{app:scaling}

\subsection{Comparing in- and out-of-distribution datasets} \label{app:comparing_features}

\begin{figure}[hbt]
    \includegraphics[width=\textwidth]{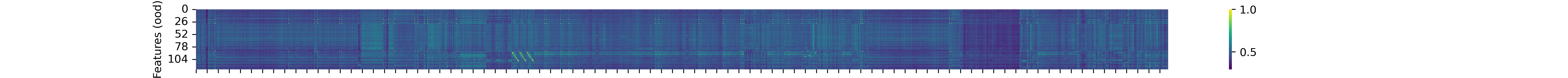}
    \caption{\textbf{Features in $\mathcal{D}_\mathrm{id}$ do not cover all features in $\mathcal{D}_\mathrm{ood}$.} Cosine similarity matrix between feature embeddings in $\mathcal{D}_\mathrm{ood}$ and $\mathcal{D}_\mathrm{id}$. For 0.752 of the ood-features, all id-features are dissimilar ( similarity$<0.8$). \textit{Note: the three little diagonal lines refer to dataset ``Kc2 Software Defects'' (OpenML ID 1063) and 3 similar datasets that contain the same features.}}
    \label{fig:cosine_similarities_ood_id}
\end{figure}

Though the number of features and samples in datasets is comparable across $\mathcal{D}_\mathrm{ood}$ and $\mathcal{D}_\mathrm{id}$ (see  Figure \ref{fig:data}), the small size of $\mathcal{D}_\mathrm{id}$ means it does not cover $\mathcal{D}_\mathrm{ood}$'s features. We plot the cosine similarities between all feature embeddings of $\mathcal{D}_\mathrm{ood}$ and $\mathcal{D}_\mathrm{id}$ in Figure \ref{fig:cosine_similarities_ood_id}. Though there is some semantic overlap in features, about three quarters of the features in $\mathcal{D}_\mathrm{ood}$ have no similar feature in $\mathcal{D}_\mathrm{id}$.

\begin{figure}[hbt]
    \centering
    \begin{subfigure}[b]{0.33\textwidth}
    \includegraphics[width=\textwidth]{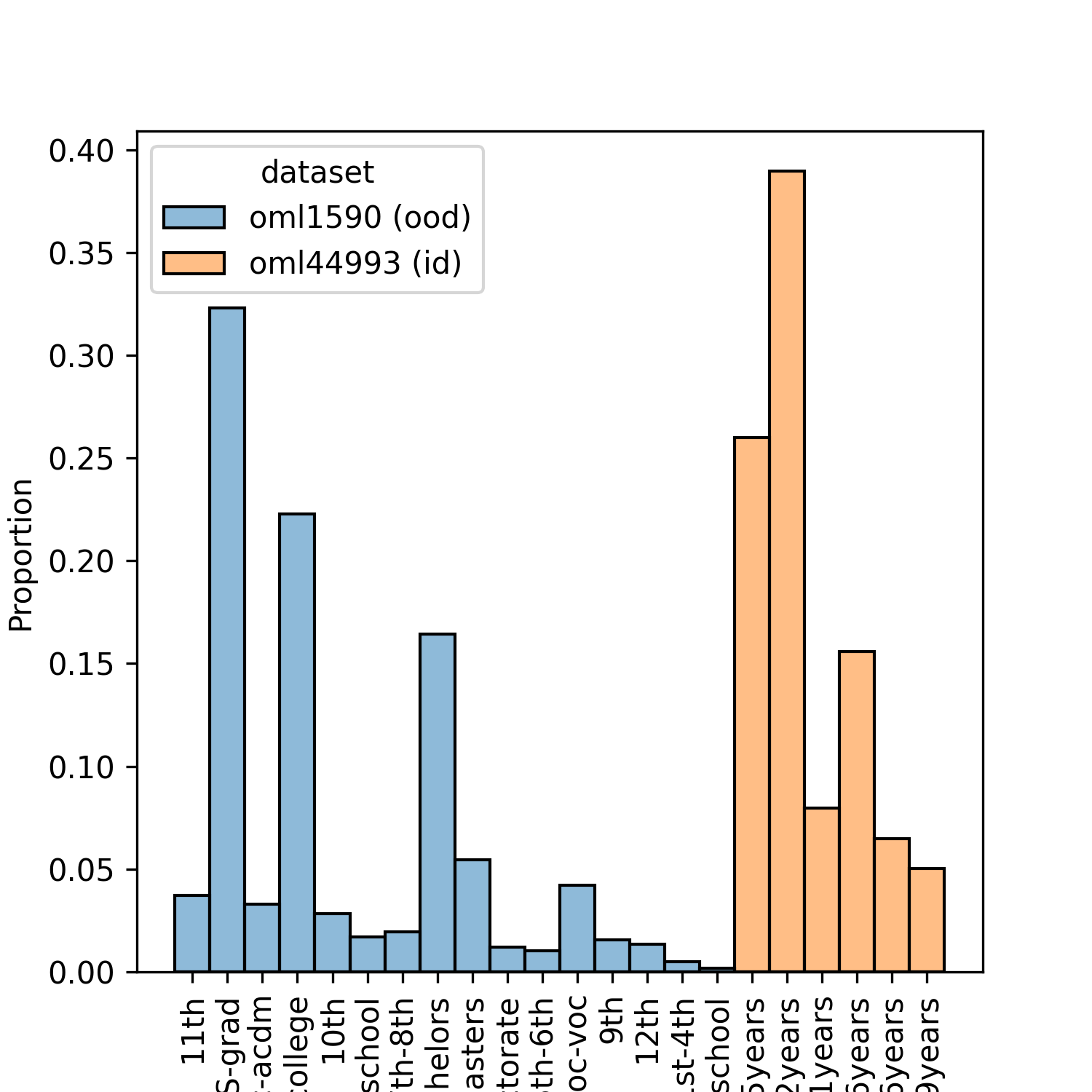}
    \caption{Education}
    \end{subfigure}
    \begin{subfigure}[b]{0.33\textwidth}
    \includegraphics[width=\textwidth]{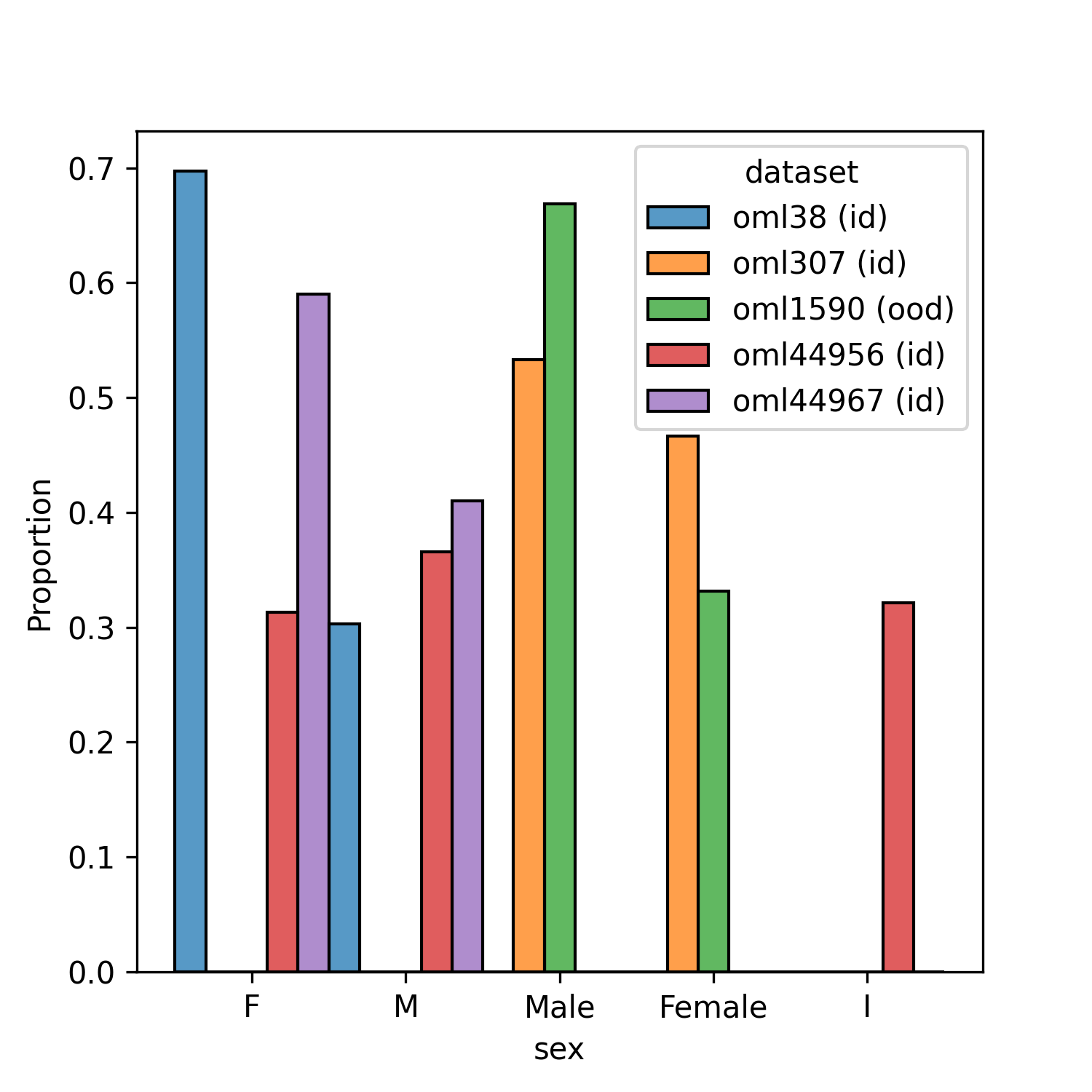}
    \caption{Sex}
    \end{subfigure}
    \begin{subfigure}[b]{0.33\textwidth}
    \includegraphics[width=\textwidth]{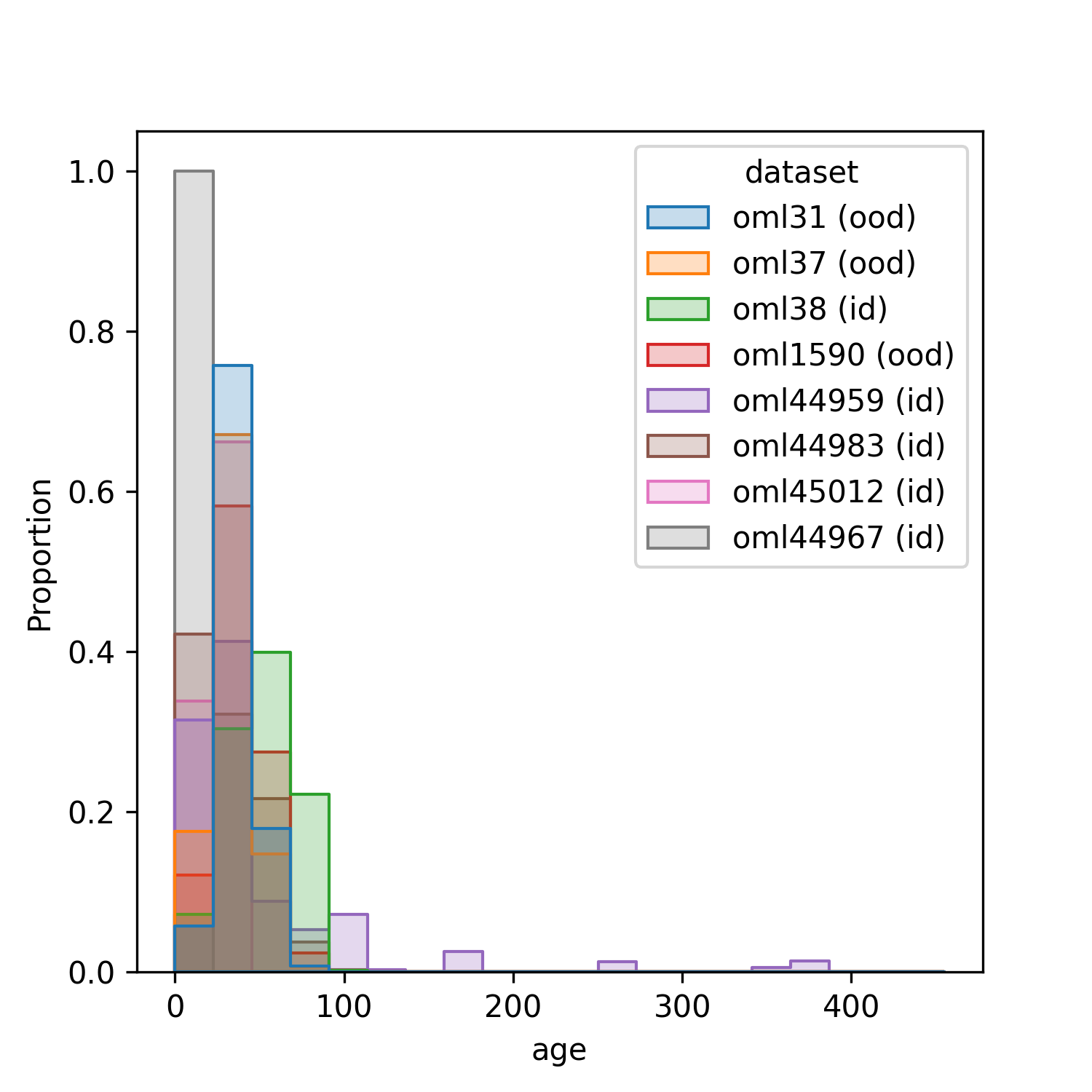}
    \caption{Age}
    \end{subfigure}~
    \caption{\textbf{Examples of feature ``matches''.} Distributions and categories can vary significantly: in (a) annotation is different, (b) there is a significant variability in female/male proportion over datasets, and in (c) there are significant differences in distribution, with one of the datasets having a vastly different scale (where age does not refer to human age).}
    \label{fig:compare_features_ood_iid}
\end{figure}

Even when similar \textit{feature names} appear in both datasets, trivially these may still pertain to vastly different features. Many datasets contain a column ``class'', yet these have vary different meaning and categories. Similarly, marginals (and joint) distribution can be significantly shifted---see Fig \ref{fig:compare_features_ood_iid}

\subsection{Using other metadatasets} \label{app:other_sources}
\begin{figure}[hbt]
    \centering
    \includegraphics[width=0.8\textwidth]{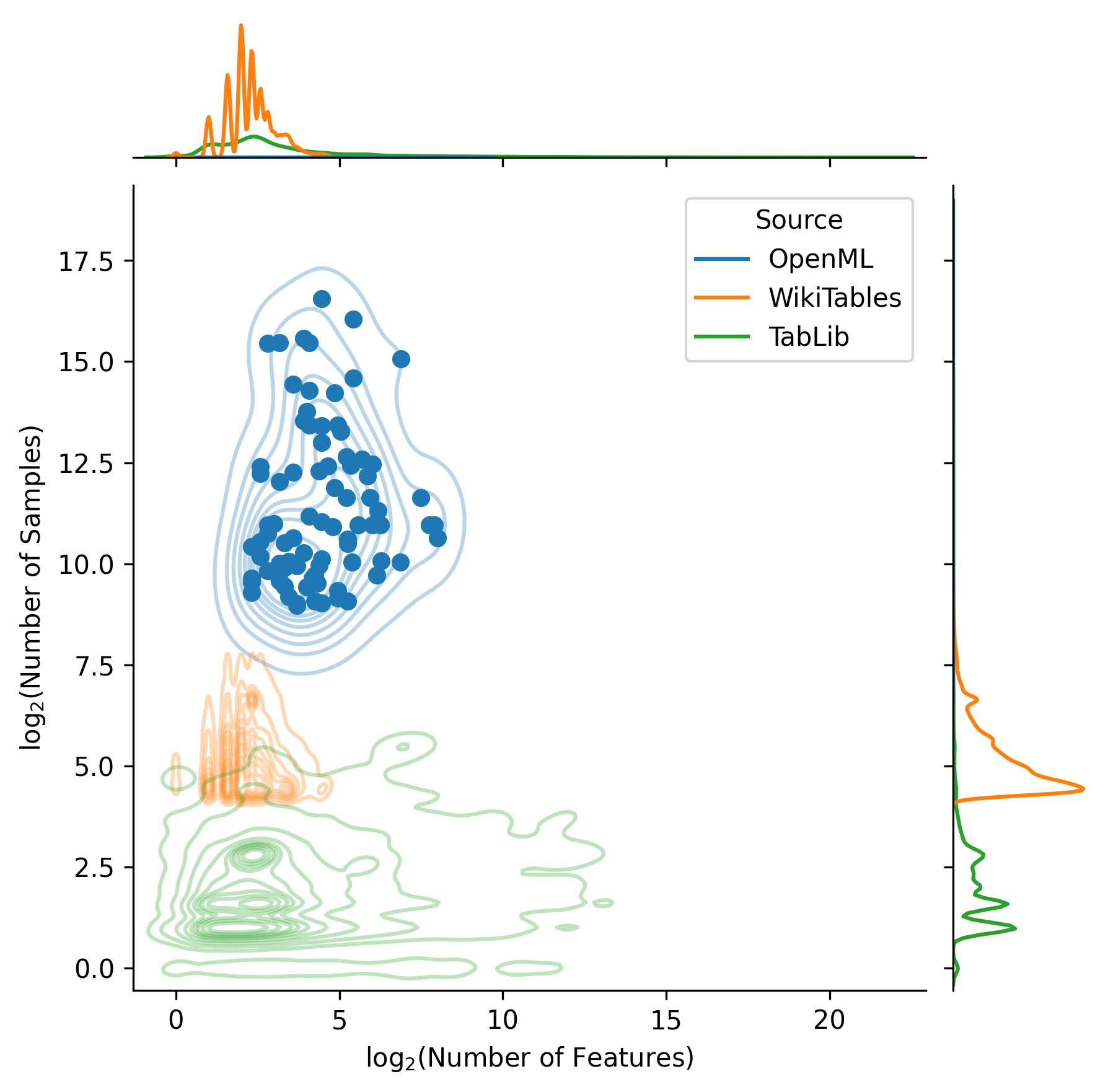}
    \caption{\textbf{Large online metadatasets have significantly different characteristics than ML tabular datasets.} Plotting different sources' number of samples and features per table. \textit{Note: double log-scale.}}
    \label{fig:compare_sources}
\end{figure}

In Figure \ref{fig:compare_sources} we compare the number of features and samples in OpenML, WikiTables, and Tablib. We observe that the latter two have significantly different characteristics. This is to be expected---many online datasets consist of informational tables (with few samples and features). TabLib consists of a large number of datasets with many features. This is likely due to a good amount of their data being sourced from Github, which may contain wide SQL and other database structures, which do not resemble typical tabular ML tasks.

\begin{figure}[hbt]
    \centering
    \begin{subfigure}[b]{0.5\textwidth}
    \includegraphics[width=\textwidth]{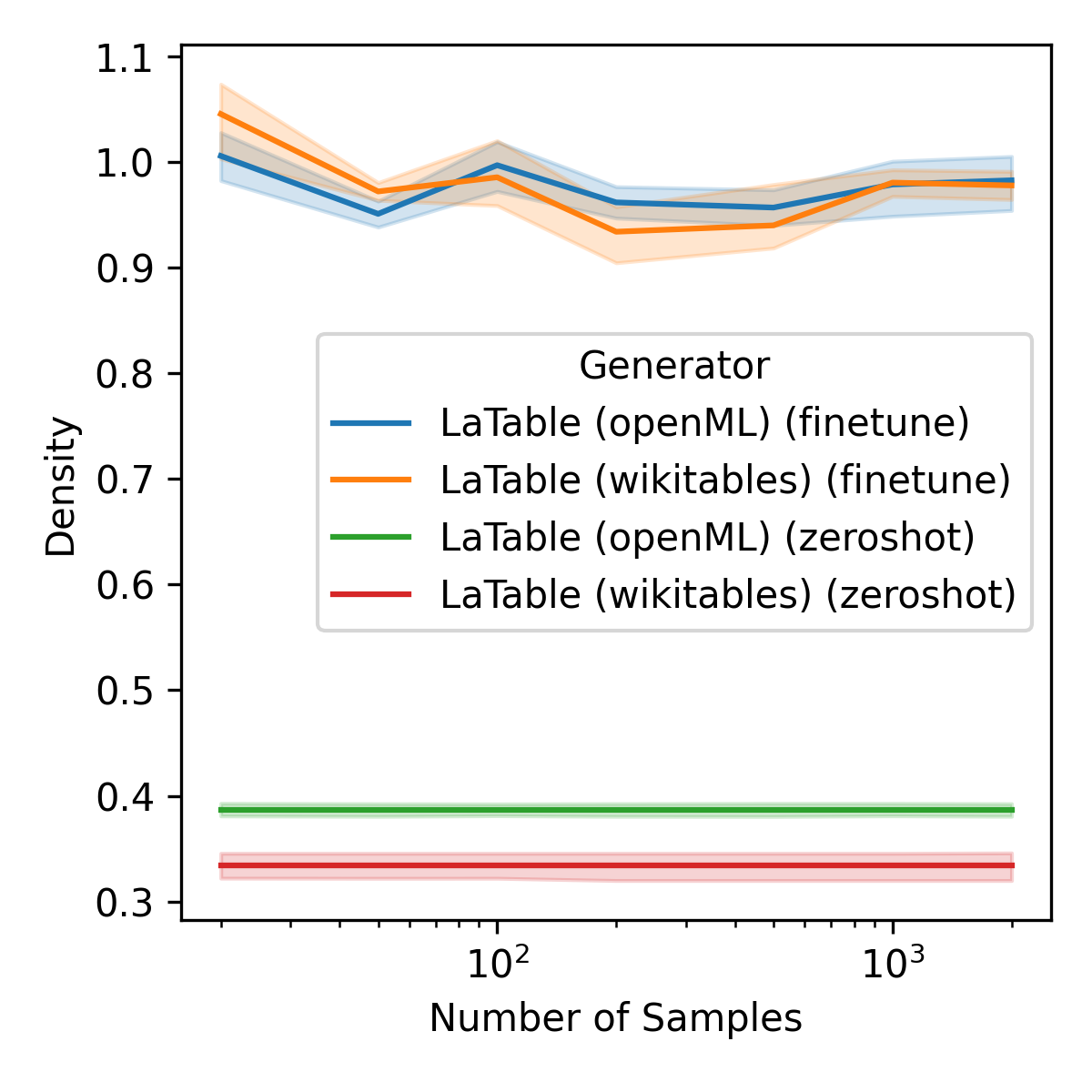}
    \caption{Density}
    \end{subfigure}~
    \begin{subfigure}[b]{0.5\textwidth}
    \includegraphics[width=\textwidth]{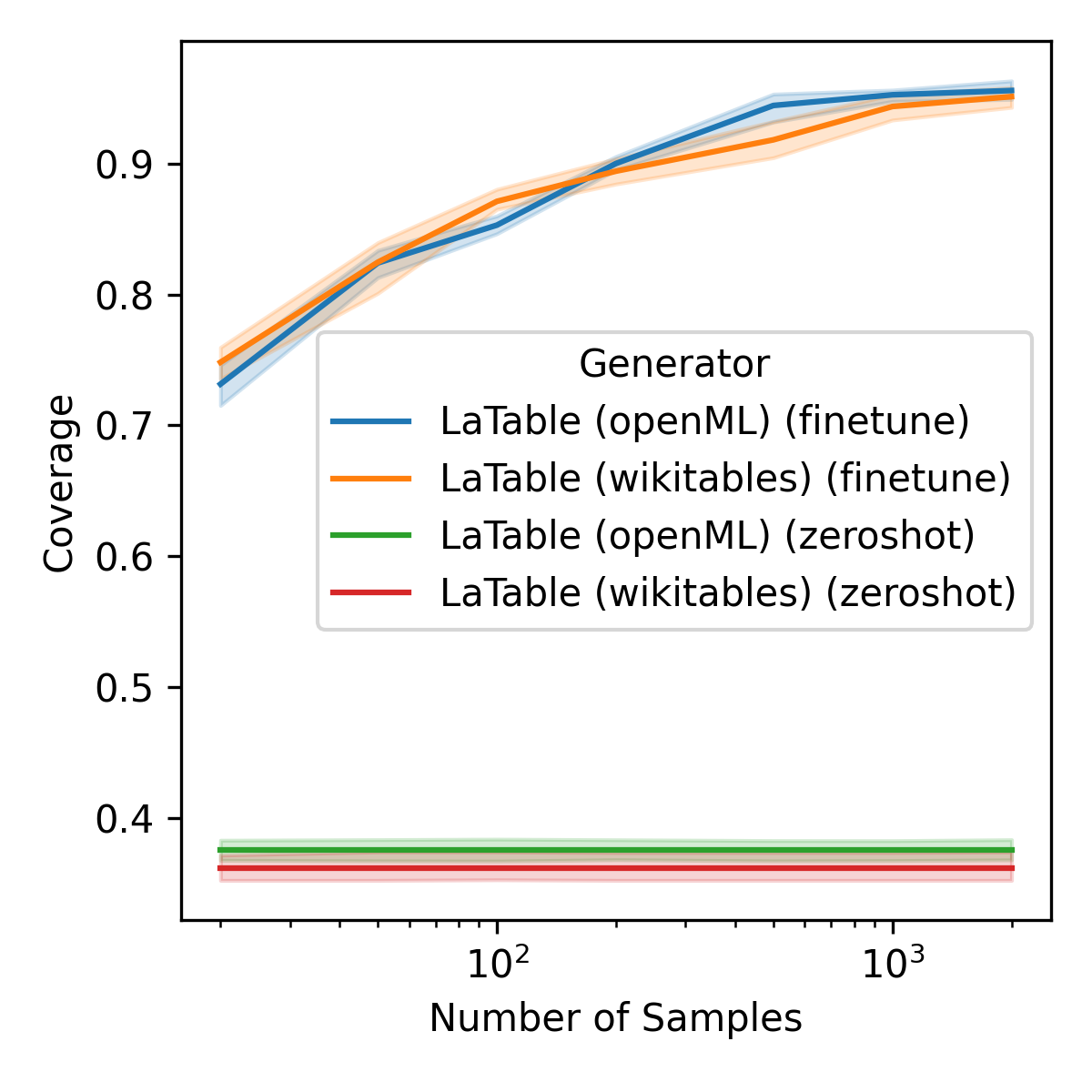}
    \caption{Coverage}
    \end{subfigure}
    \caption{\textbf{WikiTables training does not help out-of-distribution generation.} Repeating Section \ref{sec:ood}'s experiments for a model trained on WikiTables (vs. OpenML as before). We observe no significant different in finetuned performance, but significant degradation in zero-shot generation.}
    \label{fig:wikitables}
\end{figure}

In Figure \ref{fig:wikitables} we show that indeed, training on WikiTables does not yield better out-of-distribution performance. Training on WikiTables for 100 epochs took 24 days on a single GPU. 

\end{document}